# Visual anomaly detection in video by variational autoencoder

Faraz Waseem (yahoo) , Rafael Perez Martinez(Stanford University), Chris Wu  (Stanford University)

## Abstract

Video anomalies detection is the intersection of anomaly detection and visual intelligence. It has commercial applications in surveillance, security, self-driving cars and crop monitoring.  Videos can capture a variety of anomalies. Due to efforts needed to label training data, unsupervised approaches to train anomaly detection models for videos is more practical  An autoencoder is a neural network that is trained to recreate its input using latent representation of input also called a bottleneck layer. Variational autoencoder uses distribution  (mean and variance) as compared to latent vector as bottleneck layer and can have better regularization effect. In this paper we have demonstrated comparison between performance of convolutional LSTM versus a variation convolutional LSTM autoencoder  https://github.com/rafapm/CS_230/.

## 1.  Introduction

We demonstrated an unsupervised deep learning model to detect video anomalies for smart surveillance. In the context of our work, we define video anomaly detection as novel objects entering the frame of view, or objects with unusual trajectories and motion patterns. A formal definition was given in Saligrama *et al.* [1], where they defined video anomalies as " the occurrence of unusual appearance or motion attributes or the occurrence of usual appearance or motion attributes in unusual *locations or times."* Some examples include but are not limited to unattended/abandoned items for long periods of time, car accidents, a person biking through a crowd of pedestrians, or people committing crimes.

In particular, we are focusing on single-scene video anomaly detection since it has the most common use in real-world applications (e.g., surveillance camera monitoring of one location for extended periods), and it is also the most common use-case in video anomaly detection. For such applications, it is more time-efficient to have a computer do this task in comparison to a person since there is nothing interesting going on for long periods of time. In fact, this is the driving force behind modern intelligent video surveillance systems. By using computer vision,  it will not only increase the efficiency of video monitoring but it will also reduce the burden on live monitoring from humans. Although this problem has been studied thoroughly in previous works [2], it is far from being fully realized since the difficulty lies in modeling anomaly events as well as handling the sparsity of events in data sets. Furthermore, not much work has been done previously on generative approaches for video anomaly detection. We explore two versions of a variational autoencoder approach for anomaly detection in this project.

## 2.  Background

Video anomalies can be thought of as the appearance of unusual objects or movements in video clips. Video anomalies are scene dependent. For example, in the pedestrian street in university town,  the presence of cars can be thought of as an anomaly because the same car is normal in city boulevards.  Normal video is needed to express a variety of objects and movements

that may occur in a particular scene. For example in video below from UCSD

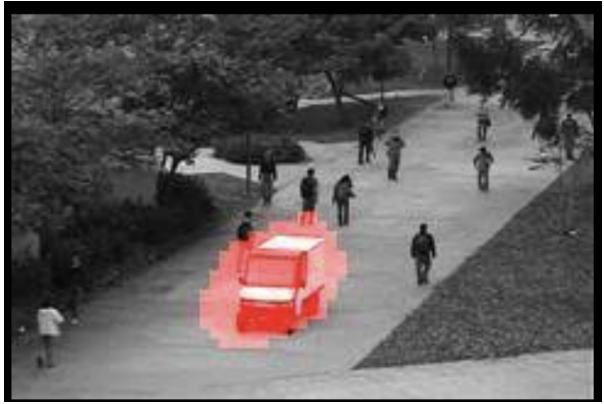

This paper is focused on single science anomaly detection because it is more common for surveillance use cases and most well published in the area of video anomaly detection research. Bharathkumar Ramachandra et al has [8] has defined following classes of anomalies

Appearance-only Anomalies

These anomalies can be thought of as unusual object appearance in a scene like a golf cart in a pedestrian scene.

Short-term Motion-only Anomalies

Something like a cycle starts moving left to right instead of forward or backward.

Long-term Trajectory Anomalies

These anomalies can be thought of as unusual object trajectory in a scene

Group Anomalies

Group anomalies can be thought of as the strange behavior of multiple objects in a group. For example if the majority in the scene are wearing a basketball cap that is anomalous unless it is a stadium while it is ok for few wearing a basketball cap in public places.

Time-of-Day Anomalies

These anomalies can be thought of as contextual to time. For example, it is normal to observe traffic outside school during pickup or drop-out time but not during evening.

These are different approaches to handle anomalies like distance measure, clustering analysis, reconstruction error, probabilistic and heuristic threshold. Heuristic thresholds are the most classic way of thinking about anomalies and require subject matter expertise in the domain. Probabilistic approaches are based on the fact that anomalies are rare events. Some examples of probabilistic approaches are Gaussian mixture etc[13] , probabilistic graphical models using features like spatio-temporal gradients [14] and optical flow fields. Cluster based approaches assume that anomalous points either don't belong to any established cluster or form their own micro clusters. Reconstruction based approaches like autoencoder are very effective for time series patterns and are based on reconstruction error A threshold learned from normal training data is used to differentiate between anomalous patterns and regular patterns. The distance based approaches like One class SVM [15] involves using training data to create a model of normality and measuring deviation from normality to determine normality score. Distance based approaches can be combined with deep learning approaches [15] where deep learning is used for feature extraction. Mixture of Gaussian models has also been used along similar lines[16] The research work we have described in this paper is focused on the reconstruction error approach.

Reconstruction based approaches encode input video from high dimensional space to low dimensional space and reconstruct video from this compact representation also called bottleneck back to original high dimensional space. It is based on the assumption that out of distributions like anomalies are harder to reconstruct. The reconstruction error can be calibrated after analyzing distribution of errors. Almost all reconstruction based approaches use deep learning and are based on either deep generative models[17] or [18] or auto-encoders [20].

In [10], the authors used convolutional auto-encoder to reconstruct training video snippets and applied pixel-wise L2 loss.

In [18] the authors built on architecture of [10] but applied convolutional LSTM to preserve temporal order of frames. Our solution is based on similar architecture but instead of using a single vector as a bottleneck layer we are using distribution (mean and variance) for richer representational power and have better regularization.

For GAN based models, the general idea is to train [21] a generator and discriminator network to differentiate between real and fake videos. As GAN (BIGAN actually ) is trained on normal training data, it will not be able to generate abnormal events and we can mark that as an anomaly.

## 3. System Overview

### 3 Dataset Description:

There are currently several datasets available for single-scene video anomaly detection. These datasets include UCSD Ped1 & Ped2 [3], CUHK Avenue [4], Subway [5], UMN [6], and Street Scene [7]. For our project, we decided to use the UCSD dataset. We chose this dataset since it has been used extensively [8, 9], giving us reference points in terms of the accuracy that can be obtained from other works in literature.

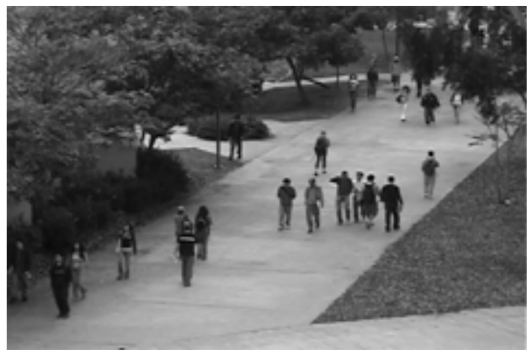

UCSD Ped1

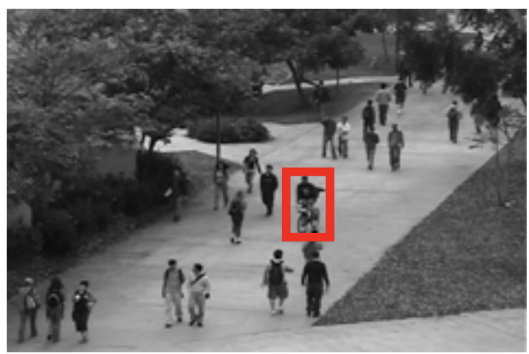

Biker

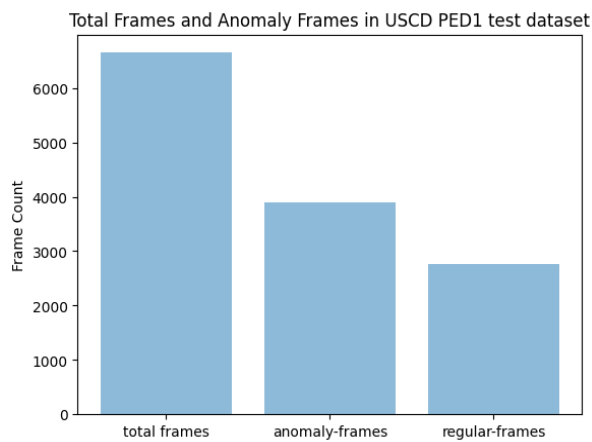

There are important points to note in this data set. The first being that the crowd density does reach a high point that leads to severe occlusions. The second being that the anomalies in both subsets are considered to be any non-pedestrian object such as bikers, skaters, carts, and wheelchairs. Any unusual motion from the pedestrians is also considered anomalous such as walking in a different pattern from what is considered normal (e.g., people walking across a walkway or at the grass).

The subset Ped1 consists of 34 training videos and 36 testing videos, all of which have a low resolution of 158 x 238 pixels. In the training videos, only normal pedestrian activity is shown. However, in the test dataset, each video clip contains at least some frames with anomalous activity. Notably, as shown in the bar chart of frame counts, the majority of frames in the test video clips contain anomalous activity (such as a biker within the frame for an extended period of time). This is not as ideal as having a more realistic distribution of anomalies in the test dataset since we expect surveillance video to be mostly normal.

## Architectures:

Description of layers

Layer Normalization :

In layer normalization input value for different neurons in the same layer is normalized across for a single training example as compared to batch normalization where input value for the same neuron is normalized for the mini batch. Batch normalization is sensitive to mini batch size and may not work if batch size is small.

TimeDistributed layer:

Time distributed layer enables us to apply the same dense layer to each temporal slice of the input . It is used when input is chronological like video frames to detect movement, actions etc. It is used when we try to model a sequence as input instead of a single frame.

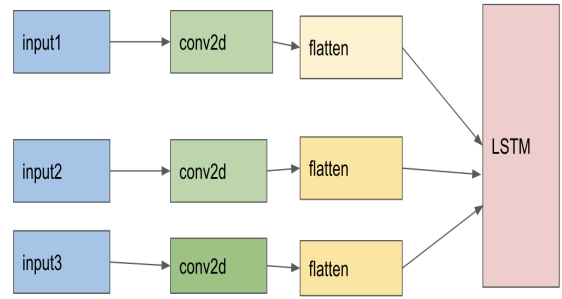

Figure Time Distributed Layer:

In the above diagram the conv2d and flatten are the same (sharing weight) for all images in sequence and this structure is shown conceptually only.

In keras we can do

model.add(TimeDistributed(Conv2D(24, activation= 'relu'), input_shape=(10, 512, 512, 3) ))

@ConvLSTM2d layer:

It is a Recurrent layer, just like the LSTM, but internal matrix multiplications are exchanged with convolution operations.It keeps input dimensions like 2d in our case.

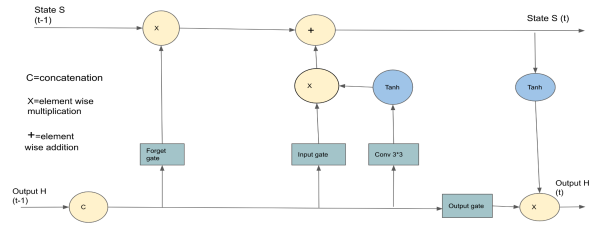

The LSTM cell input is a set of data over time, that is, a 3D tensor with shape (samples, time_steps, features). The Convolution layer input is a set of images as a 4D tensor with shape (samples, channels, rows, cols). The input of a ConvLSTM is a set of images over time as a 5D tensor with shape (samples, time_steps, channels, rows, cols).

The ConvLSTM layer output is a combination of a Convolution and a LSTM output. Just like the LSTM, if return_sequences = True, then it returns a sequence as a 5D tensor with shape (samples, time_steps, filters, rows, cols). On the other hand, if return_sequences = False, then it returns only the last value of the sequence as a 4D tensor with shape (samples, filters, rows, cols).

### Deconvolution layer (Conv2dTranspose)

Deconvolution layers are also called transposed convolution for upsampling and are usually used in autoencoders and GANs for image reconstruction. They make the output image larger in size (increase in dimension) by using a kernel which is learned using back-propagation similar to the kernel used in convolution layers.

### Custom Lambda layer:

In the VAE, we used a custom layer to create a representation of the latent distribution mean and standard deviation. Custom layer is used for the reparameterization trick.

### Network initialization:

In terms of initialization and training for all our architectures, we used the Xavier algorithm to prevent the signal from decaying to zero or "exploding" to a large value. To optimize our first architecture (LSTM Convolutional Autoencoder) we used a mean squared error loss function, with Adam optimizer. For the second and third model versions (LSTM Convolutional Variational Autoencoder), we used a special VAE loss function described later in this paper.

### 1st architecture: LSTM Convolutional Autoencoder

As a starting point, we first implemented a convolutional LSTM Convolutional autoencoder in Keras similar to the one implemented in [10]. For the autoencoder, we first defined the encoder and the decoder. The encoder takes as input a sequence of frames in chronological order. It was also divided into two parts, the spatial encoder, and the temporal encoder, where the output of the spatial encoder is the input of the temporal encoder to provide motion encoding. The purpose of the decode is to reconstruct the video sequence by mirroring the encoder.

### Architecture diagram of our baseline LSTM Convolutional Autoencoder:

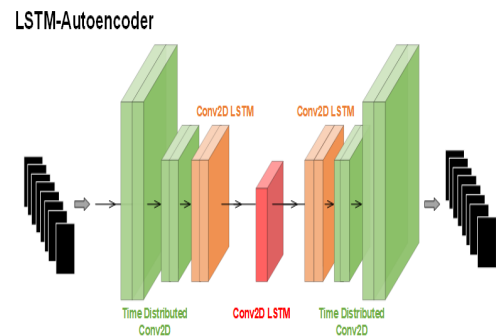

```
Model: "sequential"
_________________________________________________________________
Layer (type)                 Output Shape              Param #
=================================================================
time_distributed (TimeDistri (None, 10, 64, 64, 128)   15616
_________________________________________________________________
layer_normalization (LayerNo (None, 10, 64, 64, 128)   256
_________________________________________________________________
time_distributed_1 (TimeDist (None, 10, 32, 32, 64)    204864
_________________________________________________________________
layer_normalization_1 (Layer (None, 10, 32, 32, 64)    128
_________________________________________________________________
conv_lst_m2d (ConvLSTM2D)    (None, 10, 32, 32, 64)    295168
_________________________________________________________________
layer_normalization_2 (Layer (None, 10, 32, 32, 64)    128
_________________________________________________________________
conv_lst_m2d_1 (ConvLSTM2D)  (None, 10, 32, 32, 32)    110720
_________________________________________________________________
layer_normalization_3 (Layer (None, 10, 32, 32, 32)    64
_________________________________________________________________
conv_lst_m2d_2 (ConvLSTM2D)  (None, 10, 32, 32, 64)    221440
_________________________________________________________________
layer_normalization_4 (Layer (None, 10, 32, 32, 64)    128
_________________________________________________________________
time_distributed_2 (TimeDist (None, 10, 64, 64, 64)    102464
_________________________________________________________________
layer_normalization_5 (Layer (None, 10, 64, 64, 64)    128
_________________________________________________________________
time_distributed_3 (TimeDist (None, 10, 256, 256, 128) 991360
_________________________________________________________________
layer_normalization_6 (Layer (None, 10, 256, 256, 128) 256
_________________________________________________________________
time_distributed_4 (TimeDist (None, 10, 256, 256, 1)   15489
=================================================================
Total params: 1,958,209
Trainable params: 1,958,209
Non-trainable params: 0
```

# Variational LSTM Convolutional Autoencoder

To improve our baseline model which was based on a Convolutional autoencoder, we introduced a generative dimension in our architecture. For autoencoder (AEC) the latent space is not continuous and does not allow easy extrapolation. In the case of variational autoencoder (VAE) instead of vectors of single points, it uses mean and standard deviation. This gives the VAE richer representation power and they are much better at mapping input samples to this rich latent space. On other hand, AECs have gaps in the way they construct latent vectors to represent input space [ (https://medium.com/@realityenginesai/understanding-variational-autoencoders-and-their-applications-81a4f99efc0d) ]. So instead of learning vectors as a bottleneck layer, we learned a Gaussian distribution which consists of both mean and standard deviation. The intuition behind this approach was that learning to minimize the difference between input and output video distribution in addition to just reconstruction error will be more generic and will add a regularization effect. It will focus less on minor details of input videos and will try to learn a more generic representation of normal data. One intuition is that it will make the model work relatively better with less sensitivity threshold for reconstruction error. VAE are generative models and can be used to generate videos as well.

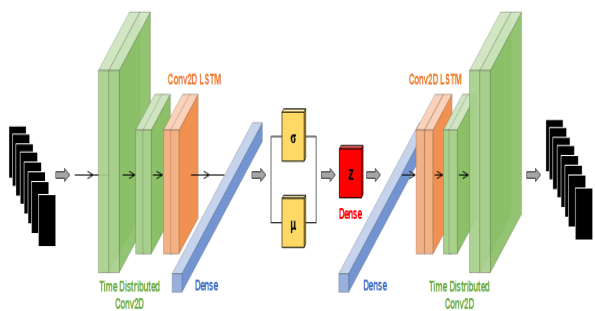

```
Model: "encoder"
________________________________________________________________________________
Layer (type)                    Output Shape         Param #     Connected to
================================================================================
encoder_input (InputLayer)      [(None, 10, 256, 256 0

time_distributed_5 (TimeDistrib (None, 10, 64, 64, 6 1664        encoder_input[0][0]

batch_normalization_8 (BatchNor (None, 10, 64, 64, 6 256         time_distributed_5[0][0]

conv_lst_m2d_2 (ConvLSTM2D)     (None, 10, 64, 64, 1 46144       batch_normalization_8[0][0]

batch_normalization_9 (BatchNor (None, 10, 64, 64, 1 64          conv_lst_m2d_2[0][0]

flatten_1 (Flatten)             (None, 655360)       0           batch_normalization_9[0][0]

dense_2 (Dense)                 (None, 32)           20971552    flatten_1[0][0]

batch_normalization_10 (BatchNo (None, 32)           128         dense_2[0][0]

latent_mu (Dense)               (None, 32)           1056        batch_normalization_10[0][0]

latent_sigma (Dense)            (None, 32)           1056        batch_normalization_10[0][0]

z (Lambda)                      (None, 32)           0           latent_mu[0][0]
                                                                 latent_sigma[0][0]
================================================================================
Total params: 21,021,920
Trainable params: 21,021,696
Non-trainable params: 224
```

```
Model: "decoder"
________________________________________________________________________________
Layer (type)                    Output Shape              Param #
================================================================================
decoder_input (InputLayer)      [(None, 32)]              0

dense_3 (Dense)                 (None, 655360)            21626880

batch_normalization_11 (Batc    (None, 655360)            2621440

reshape_1 (Reshape)             (None, 10, 64, 64, 16)    0

conv_lst_m2d_3 (ConvLSTM2D)     (None, 10, 64, 64, 16)    18496

batch_normalization_12 (Batc    (None, 10, 64, 64, 16)    64

time_distributed_6 (TimeDist    (None, 10, 256, 256, 64)  25664

batch_normalization_13 (Batc    (None, 10, 256, 256, 64)  256

time_distributed_7 (TimeDist    (None, 10, 256, 256, 1)   7745
================================================================================
Total params: 24,300,545
Trainable params: 22,989,665
Non-trainable params: 1,310,880
```

One tricky problem we encountered with variational autoencoders is how backpropagation will work with a stochastic sampling layer. The solution is the reparameterization trick which works as follows. We can compute the gradients of the sampling node with respect to the mean and log-variance vectors (both the mean and log-variance vectors are used in the sampling layer). This is represented in the following figure:

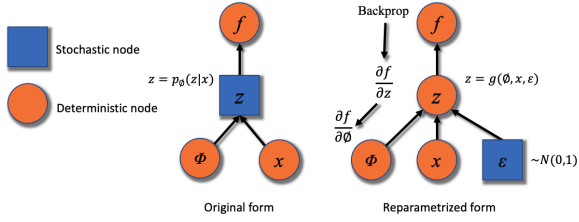

We experimented with two versions of this model architecture, by changing the loss functions. Specifically, we modified the weight of the KL-divergence term in the loss function.

The Loss function for VAE version 1 contains mean square error loss plus KL-divergence term (where Beta = 1):

$$\mathcal{L} = \mathbb{E}_{q(z|X)}[\log p(X|z)] - \beta D_{KL}[q(z|X)||p(z)]$$

The Loss function for VAE version 2 contains a Beta weight term greater than 1, which has the effect of adding a higher penalty for regularization.

$$\mathcal{L} = \mathbb{E}_{q(z|X)}[\log p(X|z)] - \beta D_{KL}[q(z|X)||p(z)]$$

We evaluated all three models on the UCSD pedestrian 1 dataset. Each model generated a regularity score output which beyond a preset threshold indicates an anomalous video frame. This regularity score was calculated as follows:

To calculate the regularity score, we first computed the reconstruction error of a pixel's intensity with value {I} at the location (x, y) in the frame {t} of the video using L2 Norm as shown in (1):

$$e(x, y, t) = \|I(x, y, t) - f_W(I(x, y, t))\|_2$$

Where fw is the learned model. The reconstruction error of a frame was then computed by summing all of the pixel-wise errors:

$$e(t) = \sum_{(x,y)} e(x, y, t).$$

We then calculate the reconstruction cost of the 10-frames sequence starting

$$\text{sequence\_reconstruction\_cost}(t) = \sum_{t'=t}^{t+10} e(t').$$

The anomaly score was then computed by scaling the reconstruction cost between 0 and 1 as shown in (4):

$$s_a(t) = \frac{\text{sequence\_reconstruction\_cost}(t) - \text{sequence\_reconstruction\_cost}(t)_{min}}{\text{sequence\_reconstruction\_cost}(t)_{max}}.$$

Finally, the regularity score was obtained by subtracting the anomaly score from 1.

$$Sr(t) = 1 - Sa(t)$$

Using the regularity scores and preset anomaly thresholds, we computed False Positive Rate, True Positive Rate (Recall), Precision, and F1 score for each model, across a range of regularity score thresholds. This allowed us to compare model performance at different sensitivity thresholds.

## 4. Case Study/Experience

For each input video clip, our models produce regularity scores such as the following (plotted with respect to video frame number):

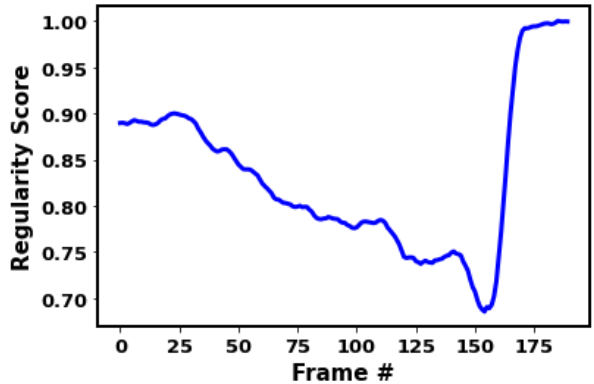

In this particular example generated by the LSTM autoencoder baseline model, a golf cart drives across the sidewalk (i.e. anomaly); therefore, we see the regularity score decrease as the golf cart drives into the foreground. Lastly, after the golf cart exits the field of view, the regularity score goes back to 1.00 (i.e. normal activity).

Comparing the F1 score vs. regulatory threshold for the three models:

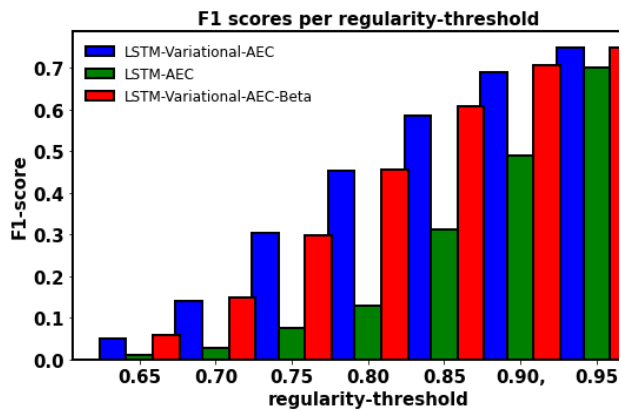

As we can see in the above plot, the LSTM-VAE-Beta model has consistently the best performance (i.e. highest F1-score) across all the regularity thresholds we tested. Both the VAEs performed similarly and were superior in terms of F1 score compared to the LSTM AE. Second, as the regularity score increases, the performance of the three models all converge to about the same F1 score. This is because as the regularity threshold approaches 1.00, the models become increasingly sensitive to any frames that might be anomalous. Since our test dataset contains a high percentage of anomalous frames (in fact, anomalous frames are the majority of frames), it makes sense that the hypersensitive model F1 scores will converge and begin to reflect the distribution of anomalous frames in the dataset. This is not practical behavior - in real use-cases, we would set the regularity threshold lower (around 0.85). In future work, it would be interesting to compare model performances on a different dataset.

We also plotted precision-recall curves for the three models:

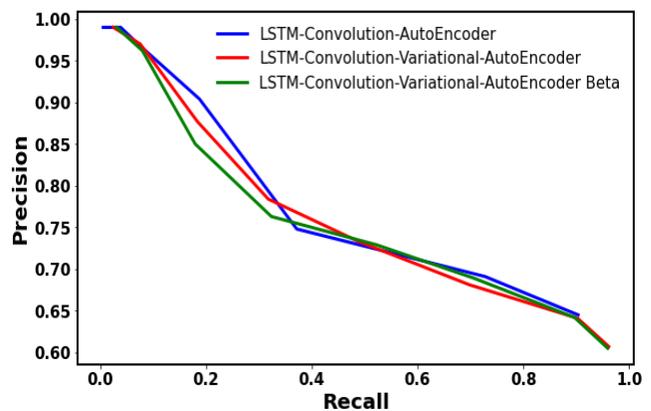

## 7. Conclusions

We found that generative models (VAEs) are better in differentiating normal and anomaly videos by learning richer representations of the input. From plotting F1 scores against different regularity scores, we can see that our performance has increased by focusing more on minimizing KL divergence between input and output videos than just minimizing the mean square error between input and output video sequences (in the LSTM AE case). We can see that variational auto-encoder outperforms vanilla LSTM-Conv Autoencoder and Beta variational autoencoder outperforms variational autoencoder slightly. This suggests that a deeper understanding (through a Gaussian distribution) of the latent space of videos has helped the VAE models with performance on test data. As for the next steps, we will do extensive hyperparameter tuning like playing with the dimension of filters, time window size (we have taken 10 as default), as well as trying different beta factors for the VAE version 2. It would also be interesting to experiment with newer datasets with existing models.

**References**

Note:

This paper is based on work done in the context of the following project.

http://cs230.stanford.edu/projects_spring_2021/reports/32.pdf